# Towards Linguistically-Aware and Language-Independent Tokenization for Large Language Models (LLMs)

Abrar Rahman
*Cognitive Computing Platform*
*Epic Systems*
Verona, WI
abrar@epic.com

Garry Bowlin
*Cognitive Computing Platform*
*Epic Systems*
Verona, WI
garry@epic.com

Binit Mohanty
*Cognitive Computing Platform*
*Epic Systems*
Verona, WI
binit@epic.com

Sean McGunigal
*Cognitive Computing Platform*
*Epic Systems*
Verona, WI
smcgunig@epic.com

*Abstract*— This paper presents a comprehensive study on the tokenization techniques employed by state-of-the-art large language models (LLMs) and their implications on the cost and availability of services across different languages, especially low resource languages. The analysis considers multiple LLMs, including GPT-4 (using cl100k_base embeddings), GPT-3 (with p50k_base embeddings), and DaVinci (employing r50k_base embeddings), as well as the widely used BERT base tokenizer. The study evaluates the tokenization variability observed across these models and investigates the challenges of linguistic representation in subword tokenization. The research underscores the importance of fostering linguistically-aware development practices, especially for languages that are traditionally under-resourced. Moreover, this paper introduces case studies that highlight the real-world implications of tokenization choices, particularly in the context of electronic health record (EHR) systems. This research aims to promote generalizable Internationalization (I18N) practices in the development of AI services in this domain and beyond, with a strong emphasis on inclusivity, particularly for languages traditionally underrepresented in AI applications.

## I. INTRODUCTION

### A. Background & Motivation

Subword tokenization techniques, such as Byte-Pair Encoding (BPE) [1] and WordPiece [2], have been instrumental in powering the subword-level tokenization approach adopted by modern large language models (LLMs). However, the prevailing tokenization methodologies raise fundamental challenges in capturing linguistically meaningful units, as they may not correspond to morphemes, the smallest units of meaning recognized by linguists [3]. The consequences of this linguistic misalignment are far-reaching, potentially resulting in suboptimal representation of languages and linguistic nuances.

Furthermore, the representation of various writing systems, encoded differently, poses technical challenges in LLM use cases, especially when aiming for global accessibility. To underscore the real-world implications, this paper reveals instances of disproportionate language tokenization costs. These challenges and disparities not only underscore the urgency for more equitable linguistic representation but also highlight the pressing need for fostering generalizable internationalization (I18N) practices in AI development that transcend language barriers and promote the inclusion of low-resource languages.

### B. Scope and Objectives

This paper's research focuses on the discrepancies in tokenization lengths across languages and tokenizers used by LLMs. The paper explores how these disparities impact various healthcare applications and clinical workflows, and the potential consequences for the quality, fairness, and accessibility of healthcare services. We aim to shed light on the ethical and technical dimensions of tokenization inequalities and future directions for the field to mitigate linguistic disparities.

## II. LITERATURE REVIEW

In early natural language processing (NLP) systems, tokenization was typically based on whitespace, where words were separated by spaces. This simplistic approach had limitations when dealing with languages that do not use spaces between words or languages with complex word forms (ex: advanced tense systems, agglutination, vowel harmony, etc). For instance, this is the default tokenizer for NLTK, a widely used NLP library [4], merely tokenizes on whitespace and gives no insight into semantically-meaningful word subunits.

### A. Subword Tokenization

- **WordPiece Tokenization**: WordPiece is an early subword tokenization method. It breaks words into smaller units, allowing the model to represent out-of-vocabulary words by combining smaller subword units [3]. However, it does not necessarily split words into semantically meaningful subunits.

- **Byte-Pair Encoding (BPE)**: Originally developed for data compression [1], this has become a widely adopted subword tokenization method in modern NLP models. BPE segments text into subword units based on character-level patterns. It dynamically generates a vocabulary of subword tokens, allowing the model to handle out-of-vocabulary words and represent morphological variations effectively.

Note that WordPiece is incredibly similar in design to BPE, albeit with a different reward function. Byte-Pair Encoding (BPE) and WordPiece are subword tokenization methods used in NLP. BPE is more flexible and identifies subword units at the character level. This allows it to better handle complex languages and out-of-vocabulary words. BPE provide more fine-grained representations and is easier to implement than WordPiece. BPE is language-agnostic and can achieve smaller vocabulary sizes. In summary, BPE is often preferred for its flexibility, representations, and language support. However, WordPiece can be a good choice depending on the task and languages. The optimal method depends on the requirements of the specific NLP application.

### B. Other Tokenization Methods

- **SentencePiece**: SentencePiece is a popular alternative to WordPiece. It is an unsupervised text tokenizer and detokenizer, which uses subword units for various NLP tasks [2]. SentencePiece provides a simpler unified interface, whereas WordPiece requires separate tokenization steps. It allows flexible segmentation strategies and operates at the character level--WordPiece uses a fixed algorithm and works at the word level. SentencePiece was designed for multilingual NLP and offers text normalization, while WordPiece originated for translating rare words without normalization.

- **Unigram Language Model (ULM)**: A tokenization method that combines ideas from BPE and language modeling. It generates subword units based on a unigram language model, optimizing the tokenization scheme according to the Minimum Description Length (MDL) principle [5].

- **Subword Regularization (SWR)**: A method for subword segmentation that aims to improve the generalization ability of NLP models. [6] It combines subword units and words, potentially offering a more intuitive representation for morphemes.

- **Sentence-BERT**: Sentence-BERT is a modification of BERT designed for sentence embeddings [7]. While not a tokenization method per se, it showcases the evolution of BERT-like models for embedding entire sentences and documents.

### C. The State of the Art for Tokenizers

OpenAI's transition from p50k_base (GPT-3) to cl100k_base (GPT-4) signifies an expansion in vocabulary size and granularity [8, 9]. These changes allow the model to handle a larger, more diverse set of subword units, enhancing its ability to represent a wider range of words and linguistic variations. This expansion is crucial for improving the model's performance, particularly when working with languages that have complex word forms, rich morphologies, and diverse writing systems.

Note that the r50k_base embedding (GPT-2) is near identical to p50k_base, with every tokenization in our analysis being completely identical between the two. Thus, r50k_base is included in the token counts but excluded for the character / token calculation.

TABLE I. COMPARISON OF EMBEDDING VOCABULARY

| LLM | Embeddings Data | |
|---|---|---|
| | *Embedding* | *Vocabulary Size (tokens)* |
| BERT | bert_base | ~30,000 |
| GPT-2 | r50k_base | ~50,000 |
| GPT-3 | p50k_base | |
| GPT-4 | cl100k_base | ~100,000 |

BERT base uses a subword tokenization method that is based on WordPiece, which generates subword units without direct consideration of linguistic morphemes. In contrast, cl100k_base uses BPE.

### D. Language Systems and Technical Challenges [3]

- **Variable-Length Encoding**: UTF-8 is a variable-length encoding, meaning that different characters use a variable number of bytes for representation. While this encoding efficiently represents characters from various languages, it can make processing more complex. This complexity can lead to parsing and indexing challenges, particularly when handling multi-byte characters, which are common in some languages.

- **Multibyte Characters**: Languages with complex scripts, such as East Asian languages (e.g., Chinese, Japanese, Korean) or Arabic (written in a calligraphic style), often have characters represented by multiple bytes in UTF-8. Processing these multibyte characters requires special handling to ensure correct indexing, substring extraction, and text manipulation. This can add computational overhead and complexity to text processing tasks.

### E. Low Resource Languages

In NLP research, a "low resource language" refers to a language for which there is limited availability of linguistic resources and data for natural language processing tasks [10]. These languages typically lack comprehensive text corpora, annotated datasets, linguistic tools, and pre-trained models.

There is no objective criterion or official list categorizing a given language as "low resource". The term can encompass several factors, including the size of the L1 (first language) speaker population. Other factors that can contribute to a language being considered low resource include the availability of digital text, linguistic diversity, and the presence of linguistic features that pose challenges for NLP, such as complex morphology or scripts. While less commonly spoken languages often face resource scarcity, it is essential to consider multiple aspects beyond just speaker population when assessing a language's resource level in NLP research. An instructive example is Bengali, a language with over 300M speakers and ranked as the 6th most spoken L1 in the world, but with relatively scant NLP research until the past five years or so [10].

### III. METHODOLOGY

#### A. Data Selection

Our heuristic for selecting content for analysis has several components: it must be widely available in languages across language families, strictly adhere to consistency of content, and preferably have some kind of literary or sociopolitical significance. With these factors in mind, we selected the United Nations Universal Declaration of Human Rights (UDHR) [11] and Genesis 1:1-3 from the Old Testament of the Bible [12-17]. We selected languages that Epic, the leading EHR in the USA, is offered in abroad [18], as well as the native Bengali of one of our authors to complete our dataset with a low-resource language. We further validated our findings on the FLORES-200 dataset. This dataset, compiled and made public by Meta Research, comprises translations of Wikimedia web articles in 200 languages of 842 articles, totaling 3001 sentences [19-21].

We considered using clinical notes, but publicly-available anonymized datasets are not available across our selection of target languages. Additionally, care as delivered is very different from national health system to health system, and thus it would not represent a true apples-to-apples comparison. Thus, the UDHR and Genesis selections highlight differences for semantically-identical strings, whereas the FLORES-200 dataset is intended to be a larger corpus of representative texts in each language.

#### B. Data Analysis Methods

Given the corpora listed above, we did a straightforward calculation of the number of characters per string. We then used OpenAI's free TikToken API [9] and the BERT base tokenizer [2] to generate token counts. We then divided columns to find the characters per token, giving us a metric for tokenization density which is a proxy for cost disparities, latency issues, and resource constraints around the size of the context window. This methodology can be extended to calculate "language premiums" as defined in [3], though we prefer having the raw data density metric as opposed to the more comparative emphasis of the language premium statistic.

### IV. RESULTS & ANALYSIS

#### A. Tokenization Variability on Short Excerpts

See Appendix I.

#### B. FLORES-200 Dataset

See Appendix I.

#### C. Impacts on LLM Cost and Availability

At Epic, we are building and deploying LLM workflows at scale, both within the US and to international customers. In doing so, the resource constraints inherent in the context windows and token allocations associated with OpenAI use cases became apparent rather quickly, which was the impetus for the internal research which gave rise to this very paper. These challenges are not restricted to healthcare workflows, but these are the example problem spaces we are currently immersed in.

- **Cost Disparities**: Tokenization premiums, as seen in LLMs charging per token, result in significant cost differences for users across languages. Processing text in languages like Bengali can cost over four times more than in English. The per-character pricing approach also leads to proportional cost disparities due to variations in character lengths across languages. An InBasket generated draft response may cost more for our Dutch customers than our American ones.

- **Latency Challenges**: High tokenization lengths in certain languages lead to longer processing times and higher latency in real-time interactions. This can result in suboptimal user experiences, miscommunication, and delays. An LLM-powered medical intake bot that you call over a phone line may be significantly slower and less usable for non-English-speaking users.

- **Long Context Processing Limitations**: Transformers models struggle to process long inputs, and tokenization directly impacts input size. For languages with high tokenization density, this limitation can significantly reduce the content processed, potentially affecting performance in any and every relevant workflow. A doctor may have to wait longer to summarize the patient's chart in Lebanon than in the US.

### V. DISCUSSION

#### A. I18N and Inclusivity in AI Services

- **Inclusivity**: We emphasize the need for AI services to be inclusive and representative of linguistic and cultural diversity. It is crucial to ensure that the representation is not biased toward dominant languages or cultures. Inclusivity should extend to under-resourced languages and marginalized communities.

- **Affordability & Access**: Researchers and developers alike must work towards AI services with larger context windows that are cost-effective and accessible, especially in regions and communities with varying economic resources.

#### B. Other Ethical Considerations

- **Preservation of Linguistic Diversity**: The advancement of AI services should not inadvertently contribute to the erosion of linguistic diversity. We seek to promote and preserve linguistic diversity by accurately representing and respecting languages with smaller speaker populations and training datasets available. Moreover, languages change over time. They are organic, living things like the people who speak them. Once a tokenization set is created and deployed broadly, they will encounter blind spots as new terms enter the lexicon.

- **Transparency and Accountability**: Developers and organizations responsible for AI services must maintain transparency in their development processes and be accountable for the impact of their technologies. Internationalization guidelines and standards should be incorporated into LLM development from early stages.

## VI. Conclusion

### A. Summary of Key Findings

The analysis of tokenization inequality challenges the conventional assumption that large language models (LLMs) can seamlessly process text in various languages. It reveals that even when intentionally trained for multilingual support, tokenization inequalities persist, leading to variations in the understanding, representation, and accessibility of different languages. This is particularly relevant for low-resource language processing because it underscores the disparities in language support, information availability, and economic costs for languages with fewer resources and smaller speaker populations. As LLM-powered workflows, in the healthcare world and far beyond, continue to scale to new markets, these problems turn more and more pressing with every passing day.

### B. Future Directions and Recommendations

- **Research and Development**: Invest in research to understand and mitigate tokenization inequalities, focusing on low-resource languages, and develop novel subword tokenization methods that prioritize fairness and inclusivity.

- **Data Collection and Collaboration**: Collaborate with linguists and language experts to curate and standardize language data for low-resource languages, improving the performance of LLMs. The field ought to work towards comprehensive transparency standards for datasets used to train tokenizers.

- **Multilingual Training**: Integrate more low-resource languages into the training pipeline of LLMs to enhance their support for linguistic diversity [22]. Engage with underrepresented language-speaking communities to better understand their unique needs and ensure that LLMs are accessible and relevant to a global audience.

- **Pricing Transparency**: Seeing as these systematic differences in tokenization and the problems they cause will not be solved overnight, we should at least be transparent about pricing and performance differences. We strongly recommend publishing "model cards" to let end users and institutions alike anticipate the cost to run a model for a given workflow on their language.

## Acknowledgment

Thank you to our many wonderful colleagues at Epic who made this paper possible: our friends Jack Dahms and Matthew Wiese on the InBasket Team; Nick Krueger on the Nebula team; and of course, our R&D Lead, Seth Hain.

APPENDIX

TABLE II: Summary Statistics for UDHR / Genesis / FLORES-200

| | Dataset | char_count | cl100k_base | p50k_base | r50k_base | bert_base | char_per_cl100k_base | char_per_p50k_base | char_per_bert_base |
|---|---|---|---|---|---|---|---|---|---|
| **Arabic (Modern Standard)** | UDHR | 116 | 88 | 120 | 120 | 94 | 1.318182 | 0.966667 | 1.234043 |
| | Genesis | 254 | 214 | 279 | 279 | 128 | 1.186915888 | 0.910394265 | 1.984375 |
| | FLORES | 111590 | 78247 | 113455 | 113455 | 90101 | 1.426124963 | 0.983561765 | 1.238499018 |
| **English** | UDHR | 170 | 33 | 33 | 33 | 32 | 5.151515 | 5.151515 | 5.3125 |
| | Genesis | 250 | 55 | 55 | 55 | 55 | 4.545454545 | 4.545454545 | 4.5454545 |
| | FLORES | 126191 | 25809 | 26738 | 26738 | 25679 | 4.88941842 | 4.719537737 | 4.914171113 |
| **Bengali** | UDHR | 168 | 212 | 337 | 337 | 128 | 0.792453 | 0.498516 | 1.3125 |
| | Genesis | 183 | 225 | 359 | 359 | 132 | 0.813333333 | 0.509749304 | 1.3863636 |
| | FLORES | 124272 | 151475 | 248271 | 248272 | 89449 | 0.820412609 | 0.500549802 | 1.389305638 |
| **Chinese (Simplified)** | UDHR | 42 | 51 | 84 | 84 | 42 | 0.823529 | 0.5 | 1 |
| | Genesis | 46 | 58 | 94 | 94 | 45 | 0.793103448 | 0.489361702 | 1.0222222 |
| | FLORES | 43181 | 48832 | 83407 | 83407 | 39030 | 0.884276704 | 0.517714341 | 1.106354087 |
| **Danish** | UDHR | 162 | 57 | 67 | 67 | 62 | 2.842105 | 2.41791 | 2.612903 |
| | Genesis | 194 | 70 | 78 | 78 | 74 | 2.771428571 | 2.487179487 | 2.6216216 |
| | FLORES | 130281 | 41899 | 49752 | 49801 | 46817 | 3.109405952 | 2.618608297 | 2.782771216 |
| **Dutch** | UDHR | 187 | 53 | 70 | 70 | 71 | 3.528302 | 2.671429 | 2.633803 |
| | Genesis | 204 | 69 | 74 | 74 | 72 | 2.956521739 | 2.756756757 | 2.8333333 |
| | FLORES | 140123 | 41022 | 51282 | 51282 | 49754 | 3.415801277 | 2.732401232 | 2.816316276 |
| **Finnish** | UDHR | 175 | 68 | 74 | 74 | 67 | 2.573529 | 2.364865 | 2.61194 |
| | Genesis | 183 | 75 | 81 | 81 | 77 | 2.44 | 2.259259259 | 2.3766233 |
| | FLORES | 134948 | 51456 | 59474 | 59474 | 50533 | 2.622590174 | 2.26902512 | 2.670492549 |
| **French** | UDHR | 186 | 50 | 61 | 61 | 61 | 3.72 | 3.04918 | 3.04918 |
| | Genesis | 240 | 74 | 83 | 83 | 84 | 3.243243243 | 2.891566265 | 2.8571428 |
| | FLORES | 149788 | 40882 | 52028 | 52028 | 46372 | 3.663910768 | 2.878988237 | 3.230138877 |
| **German** | UDHR | 164 | 44 | 60 | 60 | 57 | 3.727273 | 2.733333 | 2.877193 |
| | Genesis | 202 | 60 | 72 | 72 | 71 | 3.366666667 | 2.805555556 | 2.845070423 |
| | FLORES | 147555 | 40644 | 56056 | 56056 | 49893 | 3.630425155 | 2.632278436 | 2.957428898 |
| **Norwegian (bokmal)** | UDHR | 166 | 52 | 67 | 67 | 60 | 3.192308 | 2.477612 | 2.766667 |
| | Genesis | 181 | 65 | 79 | 79 | 68 | 2.784615385 | 2.291139241 | 2.6617647 |
| | FLORES | 127887 | 40382 | 48758 | 48758 | 45955 | 3.166930811 | 2.622892654 | 2.782874551 |
| **Russian** | UDHR | 160 | 74 | 172 | 172 | 133 | 2.162162 | 0.930233 | 1.203008 |
| | Genesis | 160 | 82 | 171 | 171 | 126 | 1.951219512 | 0.935672515 | 1.2698413 |
| | FLORES | 139099 | 64429 | 149244 | 149244 | 114648 | 2.158950162 | 0.932024068 | 1.213270184 |
| **Spanish** | UDHR | 171 | 44 | 58 | 58 | 57 | 3.886364 | 2.948276 | 3 |
| | Genesis | 235 | 74 | 88 | 88 | 91 | 3.175675676 | 2.670454545 | 2.5824175 |
| | FLORES | 150230 | 39968 | 52117 | 52117 | 48128 | 3.758757 | 2.882553 | 3.121468 |
| **Swedish** | UDHR | 163 | 56 | 73 | 73 | 58 | 2.910714 | 2.232877 | 2.810345 |
| | Genesis | 170 | 66 | 82 | 82 | 66 | 2.575757576 | 2.073170732 | 2.575757576 |
| | FLORES | 127480 | 40770 | 51269 | 51316 | 44132 | 3.126808928 | 2.486492812 | 2.888606907 |
| **Tagalog** | UDHR | 192 | 66 | 81 | 81 | 71 | 2.909091 | 2.37037 | 2.704225 |
| | Genesis | 259 | 90 | 100 | 100 | 98 | 2.877777778 | 2.59 | 2.6428571 |
| | FLORES | 159100 | 53148 | 59256 | 59378 | 55784 | 2.993527508 | 2.684960173 | 2.852072279 |
| **Vietnamese** | UDHR | 215 | 134 | 191 | 191 | 72 | 1.604478 | 1.125654 | 2.986111 |
| | Genesis | 237 | 118 | 224 | 224 | 96 | 2.008474576 | 1.058035714 | 2.46875 |
| | FLORES | 132988 | 62913 | 117465 | 117465 | 51048 | 2.113839747 | 1.132150002 | 2.605155932 |